\documentclass[conference]{IEEEtran}
\IEEEoverridecommandlockouts
\usepackage{cite}
\usepackage{amsmath,amssymb,amsfonts}
\usepackage{algorithmic}
\usepackage{graphicx}
\usepackage{textcomp}
\usepackage{xcolor}
\usepackage{hyperref}

\usepackage{booktabs}
\usepackage{multirow}
\usepackage{bookmark}
\usepackage{listings}
\makeatletter
\newcommand\footnoteref[1]{\protected@xdef\@thefnmark{\ref{#1}}\@footnotemark}
\makeatother

\def\BibTeX{{\rm B\kern-.05em{\sc i\kern-.025em b}\kern-.08em
    T\kern-.1667em\lower.7ex\hbox{E}\kern-.125emX}}
\begin{document}

\title{Uncertainty-Aware Boosted Ensembling in Multi-Modal Settings}

\author{\IEEEauthorblockN{Utkarsh Sarawgi\textsuperscript{\textsection}}
\IEEEauthorblockA{\textit{MIT Media Lab} \\
\textit{Massachusetts Institute of Technology}\\
Cambridge, USA \\
utkarshs@mit.edu}
\and
\IEEEauthorblockN{Rishab Khincha\textsuperscript{\textsection}}
\IEEEauthorblockA{\textit{MIT Media Lab} \\
\textit{Massachusetts Institute of Technology}\\
\textit{BITS Pilani, K K Birla Goa Campus}\\
Bangalore, India \\
rkhincha@mit.edu}
\and
\IEEEauthorblockN{Wazeer Zulfikar\textsuperscript{\textsection}}
\IEEEauthorblockA{\textit{McGovern Institute for Brain Research} \\
\textit{Massachusetts Institute of Technology}\\
Cambridge, USA \\
wazeer@mit.edu}
\and\hspace{32mm}
\IEEEauthorblockN{Satrajit Ghosh}
\IEEEauthorblockA{\hspace{32mm}\textit{McGovern Institute for Brain Research} \\
\hspace{32mm}\textit{Massachusetts Institute of Technology}\\
\hspace{32mm}\textit{Harvard Medical School}\\
\hspace{32mm}Cambridge, USA \\
\hspace{32mm}satra@mit.edu}
\and\hspace{-10mm}
\IEEEauthorblockN{Pattie Maes}
\IEEEauthorblockA{\hspace{-10mm}\textit{MIT Media Lab} \\
\hspace{-10mm}\textit{Massachusetts Institute of Technology}\\
\hspace{-10mm}Cambridge, USA \\
\hspace{-10mm}pattie@mit.edu}
}

\maketitle
\begingroup\renewcommand\thefootnote{\textsection}
\footnotetext{Equal contribution}
\endgroup

\begin{abstract}
Reliability of machine learning (ML) systems is crucial in safety-critical applications such as healthcare, and uncertainty estimation is a widely researched method to highlight the confidence of ML systems in deployment. Sequential and parallel ensemble techniques have shown improved performance of ML systems in multi-modal settings by leveraging the feature sets together. We propose an uncertainty-aware boosting technique for multi-modal ensembling in order to focus on the data points with higher associated uncertainty estimates, rather than the ones with higher loss values. We evaluate this method on healthcare tasks related to Dementia and Parkinson's disease which involve real-world multi-modal speech and text data, wherein our method shows an improved performance. Additional analysis suggests that introducing uncertainty-awareness into the boosted ensembles decreases the overall entropy of the system, making it more robust to heteroscedasticity in the data, as well as better calibrating each of the modalities along with high quality prediction intervals. We open-source our entire codebase at \href{https://github.com/usarawgi911/Uncertainty-aware-boosting}{https://github.com/usarawgi911/Uncertainty-aware-boosting}. 
\end{abstract}



\section{Introduction}\label{sec:introoo}

Rapid developments in machine learning (ML) across a variety of tasks have advanced its deployment in real-world settings \cite{lecun2015deep}. However, recent works have shown how these models are usually overconfident at predicting probability estimates representative of the true likelihood, and can lead to confident incorrect predictions \cite{guo2017calibration}. This is particularly detrimental in real-world domains as the distribution of the observed data may shift and eventually be very different once a model is deployed in practice, leading to models exhibiting unexpectedly poor behaviour upon deployment \cite{d2020underspecification}.

As such, generating confidence intervals or uncertainty estimates along with the predictions is crucial for reliable and safe deployment of machine learning systems in safety-critical settings (such as healthcare) \cite{amodei2016concrete, varshney2017safety, kumar2019verified, thiagarajan2020building}. It is critical to understand what a model does not know when building and deploying machine learning systems to help mitigate possible risks and biases in decision making \cite{gal2016uncertainty}. This can also help in designing reliable human-assisted AI systems for improved and more transparent decision making as the human experts in the process can account for the confidence measures of the models for a final decision.


Our experience of the world is multi-modal; data tend to exist with multiple modalities such as images, audio, text, and more in tandem. Interpreting these signals together by designing models that can process and relate information from multiple sources can help leverage different feature sets together for better understanding and decision making \cite{baltruvsaitis2018multimodal}. Parallel and sequential techniques are widely used to improve performance by ensembling weak learners trained with a single data modality \cite{breiman1996bagging, freund1997decision, freund1997adaboost, friedman2000gradientboosting, chen2016xgboost}. Similarly, base learners trained with different input modalities can be ensembled together for performance improvements \cite{baltruvsaitis2018multimodal, sarawgi2020multimodal, zhang2019ensemble}.

Some works have briefly discussed uncertainty estimation in multi-modal settings \cite{sarawgi2020unified, oviatt2000designing, sen1996bayesian, hill1993strategy}. We propose a notion of uncertainty-awareness with sequentially boosted ensembling in multi-modal settings. Particularly, we design an `uncertainty-aware boosted ensemble' for a multi-modal system where each of the base learners correspond to the different modalities. The ensemble is trained in a way such that the base learners are sequentially boosted by weighing the loss with the corresponding data point's predictive uncertainty (Section~\ref{sec:definition}).

The motivation is to sequentially boost the data points for which a particular base learner is more uncertain about its prediction. Multi-modal data, in nature, can be more prone to noise in particular modalities due to various reasons (such as the stochastic data generation process at the source). With uncertainty estimation, the noisy modalities will exhibit high uncertainty with the predictions. In such situations, having the base learners pay more attention to such uncertain predictions can help design a more robust ensemble learner. This mechanism decreases the overall entropy of the multi-modal system while generating uncertainty estimates, thus making it more reliable.

We evaluate our method on multi-modal speech and text datasets on healthcare tasks related to Dementia and Parkinson's disease using different machine learning models (Neural Networks and Random Forests) and uncertainty estimation techniques (Gaussian target distribution \cite{lakshminarayanan2017simple} and Infinitesimal Jackknife method \cite{wager2014jackknife}). Our analysis shows an increased reduction in the entropy as we sequentially move from the first base learner to the last base learner of the ensemble, further demonstrating the significance of introducing uncertainty-awareness into the ensemble. The model becomes more robust to heteroscedasticity in the data while also well calibrating each of the individual modalities along with high quality prediction intervals (Section~\ref{sec:results}). To the best of our knowledge, we are the first to explore such a boosting mechanism in a multi-modal ensemble using predictive uncertainty estimates.

\subsection{Summary of Contributions:}

\begin{figure*}[htbp] 
\centering 
  \begin{minipage}{.47\textwidth}
  \centering
  \includegraphics[width=0.95\linewidth]{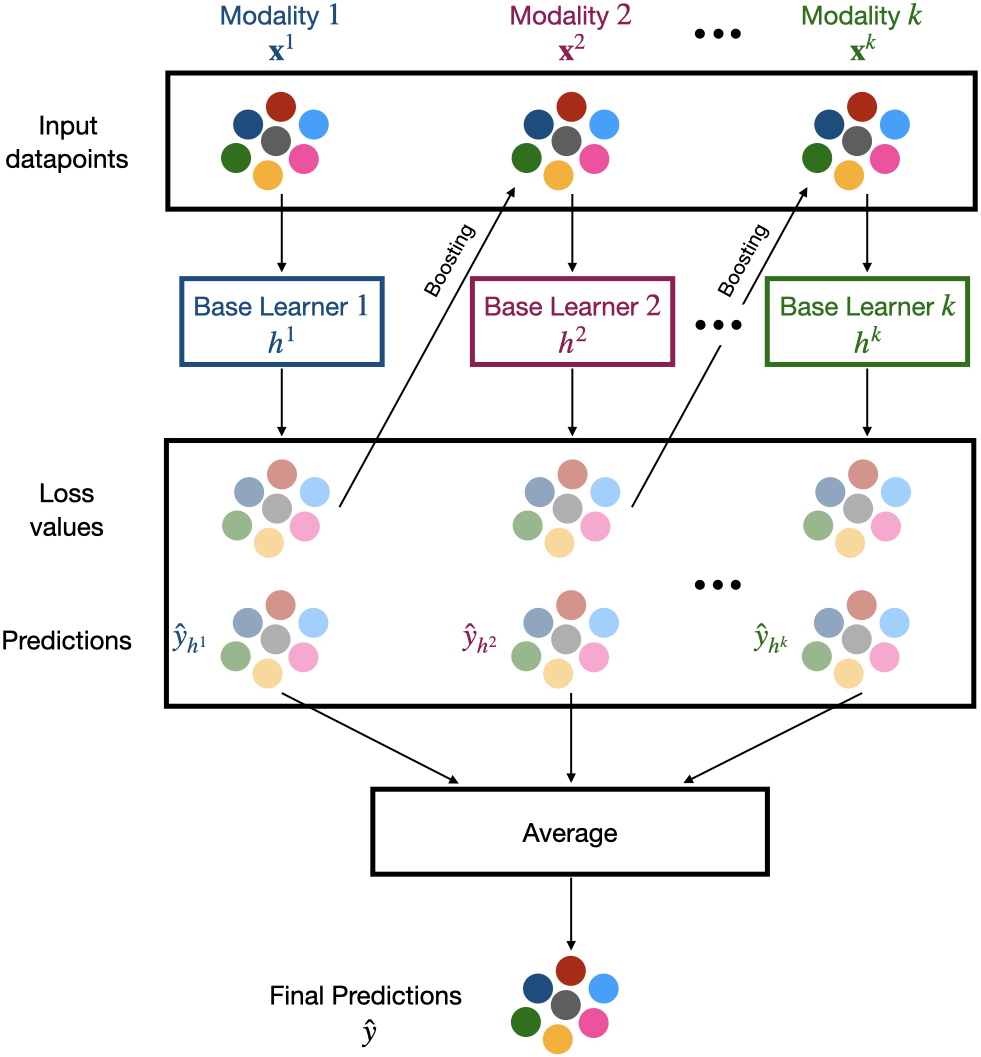}
\end{minipage}%
~~~~~~
\begin{minipage}{.47\textwidth}
\centering
  \includegraphics[width=0.95\linewidth]{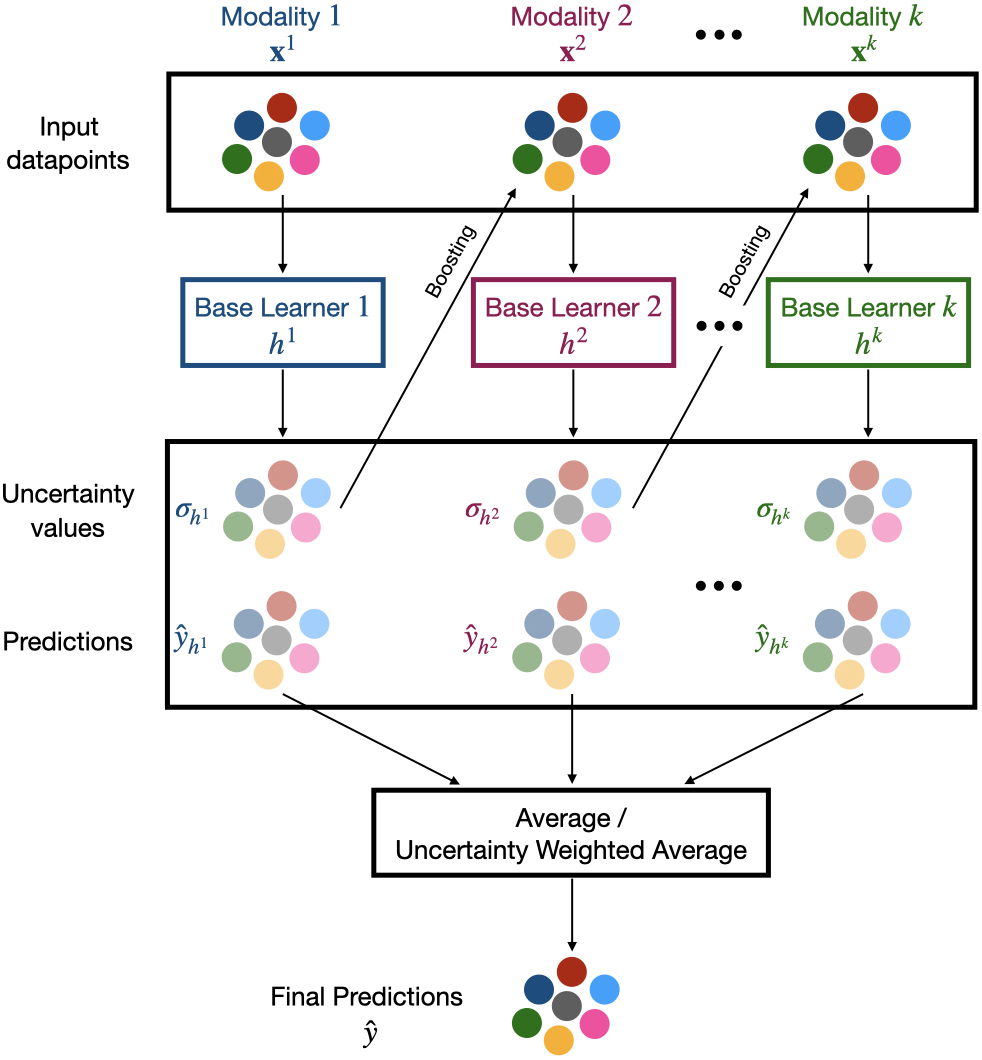}
\end{minipage}%
  \caption{Process diagrams of 1) `vanilla ensemble' (left), and 2) `uncertainty-aware (UA) ensemble' / `UA ensemble (weighted)' (right). The symbols in the diagrams follow from the notation and setup in Section \ref{sec:setup}.}\label{fig:models}
\end{figure*}

\begin{itemize}
    \item We first propose and formulate an `uncertainty-aware boosted ensemble' (Section~\ref{sec:definition}), as also depicted in the process diagram in Fig.~\ref{fig:models}.
    \item We then evaluate our method on multi-modal speech and text datasets on healthcare tasks related to Dementia and Parkinson's disease using different ML models and uncertainty estimation techniques (Section~\ref{sec:methods}).
    \item We also perform entropy, calibration, and prediction interval analyses to highlight the significance of introducing uncertainty-awareness into the ensemble (Section~\ref{sec:methods}).
\end{itemize} 

\section{Related Work}

\subsection{Uncertainty Estimation}

Numerous works have proposed a variety of both Bayesian and non-Bayesian methods to model the heteroscedasticity introduced by the stochastic data generation process for predicting the uncertainty estimates along with a model's predictions. Bayesian approximation techniques such as dropout-based VI \cite{kingma2015dropout, gal2016dropout}, expectation propagation \cite{hernandez2015probabilistic}, variational inference (VI) \cite{graves2011practical, blundell2015varinf}, deterministic VI \cite{wu2018deterministic}, neural networks as Gaussian processes \cite{lee2017deep}, approximate Bayesian ensembling \cite{pearce2020uncertainty}, and Bayesian model averaging in low-dimensional parameter subspaces \cite{izmailov2020subspace} have been shown to be quite useful in modelling the uncertainties in neural networks. Non-Bayesian approaches \cite{osband2016risk, lakshminarayanan2017simple, dusenberry2020analyzing, jain2020maximizing, sarawgi2020unified} that involve bootstrapping and ensembling multiple probabilistic neural networks have shown performances comparable to Bayesian methods with reduced computational costs and modifications to the training procedure. Additionally, there is a breadth of other theoretical, empirical, and review works on estimating predictive uncertainties with neural networks \cite{mackay1992laplace, kay1999statistics, welling2011stochastic, kendall2017uncertainties, shridhar2018uncertainty, snoek2019can, Qiu2020Quantifying}.

Uncertainty estimation in trees and random forests has been studied in the past, with multiple methods being proposed for both classification and regression tasks. NGBoost generalizes gradient boosting to probabilistic regression by treating the parameters of the conditional distribution as targets for a multiparameter boosting algorithm \cite{duan2020ngboost}. Ensemble methods have been proposed to study the uncertainty in gradient boosted models \cite{malinin2021uncertainty}. Recently, various methods such as the infinitesimal jackknife \cite{wager2014jackknife}, monte carlo based approaches \cite{coulston2016uncertainty}, and using the decision tree as a probabilistic predictor \cite{shaker2020uncertainty} have been suggested to predict the uncertainty in random forests. Ashukha et al. \cite{ashukha2020pitfalls} performed a broad study of ensembling techniques in context of uncertainty estimation.

\subsection{Ensemble techniques}
Parallel and sequential ensembling techniques have been widely studied and shown to improve performance of models in a variety of tasks \cite{breiman1996bagging, freund1997decision, freund1997adaboost, friedman2000gradientboosting, chen2016xgboost, baltruvsaitis2018multimodal, sarawgi2020multimodal, zhang2019ensemble}. Some works combine latent embeddings from different input modalities (feature fusion) and train the entire model together in a joint fashion \cite{nakamura2019r}. Boosting algorithms such as Adaptive Boost \cite{freund1997adaboost}, Gradient Boosting \cite{friedman2000gradientboosting}, and XG Boost \cite{chen2016xgboost} are common sequential ensembling techniques. These methods are often used to improve performance by ensembling weak base learners trained with a single data modality \cite{breiman1996bagging, freund1997decision, freund1997adaboost, friedman2000gradientboosting, chen2016xgboost}. Similarly in multi-modal settings, base learners trained with different input data modalities can be ensembled together to help leverage different feature sets together \cite{baltruvsaitis2018multimodal, sarawgi2020multimodal, zhang2019ensemble}.

\vspace{-5mm}
\subsection*{ }
There has been some work on incorporating uncertainty with ensembling methods. Kendall et al. \cite{kendall2018multitask} learns multiple tasks by using the uncertainty predicted as weights for the losses of each of the models, thus outperforming the individual models trained on each task. Chang et al. \cite{chang2017activebias} uses uncertainty estimates and prefers to learn the data points predicted incorrectly with higher uncertainty in different mini-batches of SGD. Some other works have addressed the advantages of incorporating uncertainty estimates in multi-modal settings \cite{sarawgi2020unified, oviatt2000designing, sen1996bayesian, hill1993strategy}.

Our work particularly designs boosted ensemble techniques that are uncertainty-aware in multi-modal settings, explores how these can be really helpful in real-world settings, and motivates future research directions.

\section{Uncertainty-Aware Boosted Ensemble}\label{sec:definition}
\subsection{Notation and Setup}\label{sec:setup}
Let $\mathbf{x}$ represent the multi-modal input feature set and $y \in \mathbb{R}$ denote the real-valued label for regression. We let $\mathbf{x}^j \in \mathbb{R}^d$ represent a set of $d$-dimensional input features for the $j^{th}$ modality, with $j = 1$ to $k$, where $k$ is the total number of modalities.

Let $\{h^j\}_{j=1}^k$ represent the corresponding base learner for the $j^{th}$ modality. The term base learner is just an abstraction for any learnt functions that maps an input to an output, for example, SVM, random forest, neural network, etc.

We subsequently have a training dataset $\{(\mathbf{x}^j_n, y_n)\}_{n=1}^N$ consisting of $N$ i.i.d. samples for the $j^{th}$ modality i.e.
\begin{equation}
    h^j : \mathbf{x}^j_n \longrightarrow y_n
\end{equation}

\subsection{Defining Uncertainty-Aware Boosted Ensemble}\label{sec:models}

We first define a `vanilla ensemble' for a fair comparison with our proposed approach. We then define our `uncertainty-aware ensemble' referred to as `UA ensemble', and its variation referred to as `UA ensemble (weighted)'. Fig.~\ref{fig:models} shows the process diagrams of vanilla ensemble, UA ensemble, and UA ensemble (weighted).

\subsubsection{Vanilla ensemble}
This makes use of loss values, i.e. mean squared error (MSE) values for regression, to weight the loss function during training while sequentially boosting across the base learners. This means that the MSE values corresponding to the predictions from the ${j}^{th}$ base learner are used to weight the loss function for the corresponding training samples while training the $(j+1)^{th}$ base learner. Then, the ensemble computes an average of the predictions $\{\hat{y}_{h^j}\}_{j=1}^k$ of all the (boosted) base learners for the final prediction $\hat{y}$.

\subsubsection{UA ensemble}
This makes use of predicted uncertainty estimates $\sigma_{h^j}$ to weight the loss function during training while sequentially boosting across the base learners. This means that the uncertainty estimates $\sigma_{h^j}$ corresponding to the predictions from the ${j}^{th}$ base learner are used to weight the loss function for the corresponding training samples while training the \hspace{5mm}$(j+1)^{th}$ base learner. Then, the ensemble computes an average of the predictions $\{\hat{y}_{h^j}\}_{j=1}^k$ of all the (boosted) base learners for the final prediction $\hat{y}$.

\subsubsection{UA ensemble (weighted)}
We experiment with a variation to the final averaging of the predictions in the UA ensemble discussed above. Here, for the final prediction $\hat{y}$, the ensemble computes a weighted average of the predictions $\{\hat{y}_{h^j}\}_{j=1}^k$ of all the (boosted) base learners, where the weights used are the inverse of the respective predicted uncertainty estimates $\sigma_{h^j}$. Equation~\eqref{eq:ua_voting} mathematically formulates this, where $\hat{y}(\mathbf{x}_n)$ is the final prediction corresponding to $n^{th}$ data point, and $k$ is the total number of individual modalities as defined in Section~\ref{sec:setup}.
\begin{align}
\begin{split}
{\hat{y}(\mathbf{x}_n)} = \frac{\sum_{j=1}^k\frac{1}{\sigma_{h^j}(\mathbf{x}_n)}\hat{y}_{h^j}(\mathbf{x}_n)}{\sum_{j=1}^k\frac{1}{\sigma_{h^j}(\mathbf{x}_n)}}\\
\label{eq:ua_voting}
\end{split}
\end{align}

\subsection*{ }
Most of the previously proposed boosting methods sequentially boost across different base learners using the same set of total input features. However, UA ensembles sequentially boost through different base learners, with each base learner corresponding to a different input modality. This is because we want to best leverage each of the modality-wise features while deriving a strong multi-modal learner using individual modality-wise base-learners together. It is important to note that unlike other boosting techniques \cite{freund1997adaboost, friedman2000gradientboosting, chen2016xgboost}, the base learners here need not be weak learners.

\section{Experiments and Results}\label{sec:methods}
We test and evaluate our proposed methods on two speech and language-based multi-modal datasets in healthcare tasks related to Dementia and Parkinson's disease. We make use of different types of machine learning models (Neural Networks and Random Forests) and uncertainty estimation techniques (Gaussian target distribution \cite{lakshminarayanan2017simple} and Infinitesimal Jackknife method \cite{wager2014jackknife}) for the two datasets.

\subsection{Datasets}\label{sec:dataset}
\subsubsection{Dementia}
We use the standardized and benchmark ADReSS (Alzheimer's Dementia Recognition through Spontaneous Speech) dataset\footnote{ADReSS dataset can be downloaded from \href{https://dementia.talkbank.org/}{https://dementia.talkbank.org/} along with an email to obtain the password for access.} \cite{luz2020alzheimer}. This dataset consists of speech samples (WAV format) and transcripts (CHA format), and their corresponding `MMSE' (Mini-Mental State Examination) scores as labels for regression. MMSE scores (ranging from 0 to 30 and widely used in clinical practice) offer a way to quantify cognitive function, as well as screening for cognitive loss by testing the individuals’ attention, recall, language, and motor skills \cite{tombaugh1992mini}.

\begin{figure*}[htbp] 
\centering 
  \begin{minipage}{.45\textwidth}
  \centering
  \includegraphics[width=0.65\linewidth]{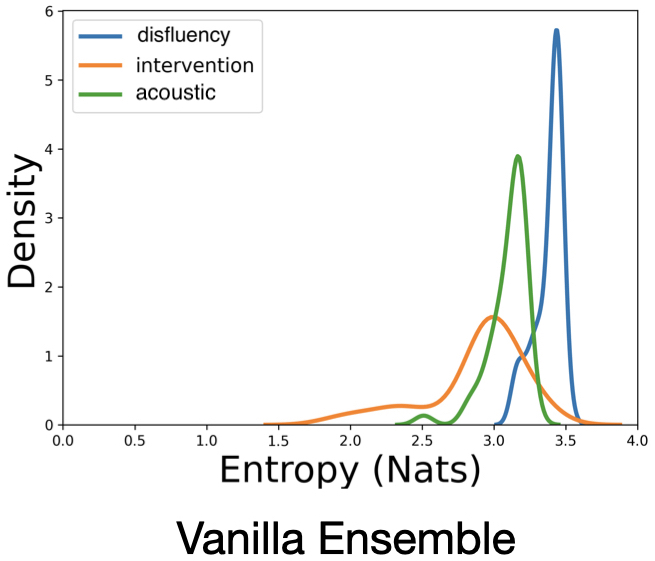}
\end{minipage}%
~~~~~
\begin{minipage}{.45\textwidth}
\centering
  \includegraphics[width=0.65\linewidth]{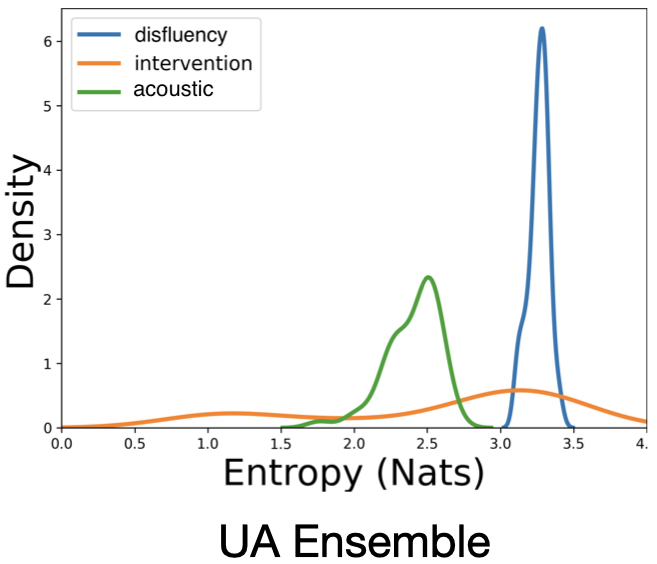}
\end{minipage}%
  \caption{Entropy analysis, using kernel density estimation plots, of the base learners in a vanilla ensemble (left) and UA ensemble (right). UA ensemble shows a decrease in the overall entropy of the system. The increased reduction in the entropy as we sequentially move from the first base learner to the last base learner of the ensemble further indicates the significance of introducing uncertainty-awareness into the ensemble (Section~\ref{sec:ad_results}).}\label{fig:entropy}
\end{figure*}

The dataset consists of 156 data points, each from a unique subject, matched for age and gender. A standardized train-test split of around 70\%-30\% (108 and 48 subjects) is provided by the dataset. We further split the train set into 80\%-20\% train-val sets. The test set was held out for all experimentation until final evaluation.

\subsubsection{Parkinson's Disease}
We use the publicly available Parkinson's Telemonitoring dataset\footnote{Parkinson's Telemonitoring dataset can be downloaded from \href{https://archive.ics.uci.edu/ml/datasets/Parkinsons+Telemonitoring}{https://archive.ics.uci.edu/ml/datasets/Parkinsons+Telemonitoring}.} \cite{tasanas2009parkinsons}. This dataset consists of a range of 16 biomedical voice measurements and their corresponding `Total UPDRS' (Unified Parkinson Disease Rating Scale) scores as labels for regression. Total UPDRS scores (ranging from 0 to 199 and widely used as a measure of severity of the Parkinson's disease (PD)) offer a way to quantify the course of PD in patients by testing the individuals’ mentation, behaviour, mood, daily-life activities, and motor examination \cite{movement2003unified}.

The dataset consists of 5,875 data points from 42 subjects with early-stage PD recruited to a six-month trial of a telemonitoring device for remote symptom progression monitoring. Since a standardized train-test split is not provided by the dataset, we use a 5-fold cross validation with consistent folds across the different methods for a fair evaluation.

\subsection{Multi-modal Feature Extraction} \label{sec:feature_eng}
\subsubsection{Dementia}
For a fair comparison with the state-of-the-art, we extract multi-modal acoustic, cognitive and linguistic features from the available speech samples and their corresponding transcripts, by using the feature engineering pipeline as developed by Sarawgi et al. \cite{sarawgi2020multimodal}. This results in three input modalities, namely `Disfluency', `Interventions', and `Acoustic' (following the same terminology as Sarawgi et al. \cite{sarawgi2020multimodal}). The details of features from each of the three modalities can be found in our Supplementary material\footnote{\label{supp}Please see \href{https://drive.google.com/file/d/1i55Lh99ojAjWWehGO52mk8UK902myFPe/view?usp=sharing}{https://tinyurl.com/16osjsix} for the Supplementary material.}.

\subsubsection{Parkinson's Disease}
The dataset consists of features related to amplitude and frequency. Accordingly, we extract two input modalities from the available data, referring to them as the `Amplitude' and `Frequency' modalities. Consequently, the list of features in the two input modalities are as below:
\begin{itemize}
\item `Amplitude': Shimmer, Shimmer(dB), Shimmer:APQ3, Shimmer:APQ5, Shimmer:APQ11, Shimmer:DDA, NHR, HNR, RPDE, DFA
\item `Frequency': Jitter(\%), Jitter(Abs), Jitter:RAP, Jitter:PPQ5, Jitter:DDP, PPE
\end{itemize}

\subsection{Model Architecture and Training}\label{sec:model_arch}

\subsubsection{Dementia}
Following from Section~\ref{sec:feature_eng}, we have three input modalities here i.e. $j = 1, 2, 3$ and $k = 3$. Now, with a training dataset $\{(\mathbf{x}^j_n, y_n)\}_{n=1}^N$ consisting of $N$ i.i.d. samples for each of the three modalities, we model the probabilistic predictive distribution $p_{h^j} (y|\mathbf{x}^j)$ using a neural network (NN) with parameters $h^j$.

For a fair comparison with the state-of-the-art, we use almost the same NN architecture as used by Sarawgi et al. \cite{sarawgi2020multimodal} for each of the three input modalities. The Disfluency and Acoustic models make use of multi-layer perceptrons (MLPs), while the Interventions model makes use of LSTM, along with regularizers. The exact model architecture, along with regularizers for each of the three models (base learners) can be found in our Supplementary material\footnoteref{supp}.

For uncertainty estimation, each of the models predicts a target distribution instead of a point estimate to account for the heteroscedasticity in data and yields predictive uncertainties along with the predicted mean value \cite{lakshminarayanan2017simple, snoek2019can, sarawgi2020unified}. The target distribution is modelled as a Gaussian distribution $p_{h^j}(y_n|\mathbf{x}^j_n)$ parameterized by the mean $\mu_{h^j}$ and the standard deviation $\sigma_{h^j}$, predicted at the final layer of the models i.e. $y_n \thicksim \mathcal{N}\left(\mu_{h^j},  \sigma_{h^j}^2\right)$. It is important to note here that the prediction $\hat{y}_{h^j}$ is the predicted mean $\mu_{h^j}$, and the predicted uncertainty estimate is the predicted standard deviation $\sigma_{h^j}$. The exact illustration of the implementation can be found in our Supplementary material\footnoteref{supp}.

Each of the three base learners is trained with their corresponding input modality features $\mathbf{x}^j$ and ground truth labels $y$ using a proper scoring rule. We optimize for the negative log-likelihood (NLL) of the joint distribution $p_{h^j}({y}_n|\mathbf{x}^j_n)$ according to the equation below~\eqref{NLL}.
\begin{align}
    \label{NLL}
    -\log \left(p_{h^j}(y_n|\mathbf{x}^j_n)\right) &= \frac{\log \left(\sigma^2_{h^j}\right)}{2} + \frac{\left(y-\mu_{h^j}\right)^2}{2\sigma_{h^j}^2}\nonumber\\
    &+ \text{constant}
\end{align}

We use the boosting methods explained in Section~\ref{sec:models} to train an ensemble with the three base learners (Disfluency, Acoustic, and Interventions). Each training run used a batch size of 32 and an Adam optimizer with a learning rate of 0.00125 to minimize the NLL.

\subsubsection{Parkinson's Disease}
Following from Section~\ref{sec:feature_eng}, we have two input modalities here i.e. $j = 1, 2$ and $k = 2$. Now, with a training dataset $\{(\mathbf{x}^j_n, y_n)\}_{n=1}^N$ consisting of $N$ i.i.d. samples for each of the two modalities, we model the probabilistic predictive distribution $p_{h^j} (y|\mathbf{x}^j)$ using a random forest (RF) regressor with parameters $h^j$. Each of the RFs makes use of 300 decision tree estimators. This was decided upon sweeping the number of decision trees, from 100 to 1000, as a hyperparameter.

The two base learners are trained with their corresponding input modality features $\mathbf{x}^j$ and ground truth labels $y$ using a mean squared error (MSE) loss. The uncertainty estimates $\sigma_{h^j}$ of each of the data point is estimated as the confidence interval using the Infinitesimal Jackknife method \cite{wager2014jackknife, wager2016forestci_2016}.

We use the boosting methods explained in Section~\ref{sec:models} to train an ensemble with the two base learners (Amplitude and Frequency).

\subsection{Results}
\label{sec:results}
\subsubsection{Dementia}\label{sec:ad_results}
For robustness, we repeat every training and test-set evaluation $5$ times and report the mean and variance of the root mean squared error (RMSE) results across the five runs. We first evaluate each of the modalities (i.e. base learners) individually and then compare them with the vanilla and uncertainty-aware ensembles. The order of sequential boosting for the propagation of the uncertainties is chosen in the order of the test set performance of the individual modalities. We observe that the uncertainty-aware ensembles perform better than the vanilla ensemble and the individual modalities (Table~\ref{tab:res1}).


\begin{table}[htbp]
\caption{Comparison of individual modalities i.e. base learners and ensemble methods on test set results of the ADReSS dataset.}
\begin{center}
\begin{tabular}{lc}
  \toprule
  \bfseries Model & \bfseries RMSE\\
  \midrule
  Disfluency & 5.71 $\pm$ 0.39\\
    Interventions & 6.41 $\pm$ 0.53\\
  Acoustic & 6.66 $\pm$ 0.30\\
  Vanilla Ensemble & 5.17 $\pm$ 0.27\\
  \textbf{UA Ensemble} & \textbf{5.05 $\pm$ 0.53}\\
  \textbf{UA Ensemble (weighted)} & \textbf{4.96 $\pm$ 0.49}\\
  \bottomrule
\end{tabular}
\label{tab:res1}
\end{center}
\end{table}

We also compare our uncertainty-aware ensemble methods with current state-of-the-art results on the ADReSS test set. Table~\ref{tab:res2} shows that the best of 5 runs of UA Ensemble is competitive, and that of the UA Ensemble (weighted) outperforms other methods. 

\begin{table}[htbp]
\caption{Comparison of uncertainty-aware ensemble methods with state-of-the-art results on the ADReSS test set.}
\begin{center}
\begin{tabular}{lc}
\toprule
\bfseries Model & \bfseries RMSE\\
\midrule
Pappagari et al. \cite{pappagariusing} & 5.37\\
Luz et al. \cite{luz2020alzheimer}   & 5.20\\
Sarawgi et al. \cite{sarawgi2020multimodal} & 4.60\\
Searle et al. \cite{searle2020comparing} & 4.58\\
Balagopalan et al. \cite{balagopalan2020bert} & 4.56\\
Rohanian et al. \cite{rohanianmulti} & 4.54\\
Sarawgi et al. \cite{sarawgi2020unified} & 4.37\\
\textbf{UA Ensemble} & \textbf{4.35}\\
\textbf{UA Ensemble (weighted)} & \textbf{3.93}\\
\bottomrule  
\end{tabular}
\label{tab:res2}
\end{center}
\end{table}

The use of uncertainty awareness can improve the robustness of the ensemble with uncertain data points (i.e subjects). To highlight this, we evaluate the entropy of the base learners in the ensemble methods while sequentially boosting in the vanilla ensemble and the UA ensemble. Upon comparison, we see a decrease in the overall entropy of the system when the ensemble is uncertainty-aware (Fig.~\ref{fig:entropy}). The increased reduction in the entropy, as we sequentially move from the first base learner to the last base learner of the ensemble, further indicates the significance of introducing uncertainty-awareness into the ensemble.

\begin{table*}[htbp]
\caption{5-times repeated test set results of Mean Prediction Interval Width (MPIW) and Prediction Interval Coverage Probability (PICP) for the ensemble techniques on the ADReSS dataset. We report PICP results with the prediction interval ($\Delta$) equal to 1, 2, and 3 times the standard deviation (i.e. $1\sigma$, $2\sigma$, and $3\sigma$). The uncertainty-aware boosting results in tighter bounds for the confidence intervals, along with higher PICP values, and high quality prediction intervals as desired (Section~\ref{sec:ad_results}).}
\begin{center}
\begin{tabular}{lccccc}
    \toprule
    \multirow{2}{*}{Model}     & \multirow{2}{*}{Modality}   & 
    \multirow{2}{*}{MPIW}   & 
    \multicolumn{3}{c}{PICP ($\%$)} \\
    \cmidrule(r){4-6} 
    &   &   & $\Delta = 1\sigma$ & $\Delta = 2\sigma$ & $\Delta = 3\sigma$\\
    \toprule
    \multirow{3}{*}{Vanilla Ensemble}
    & Disfluency & 4.47 $\pm$ 0.39 & 61.66 $\pm$ 8.29 & 95.83 $\pm$ 2.63 & 97.50 $\pm$ 0.83 \\
    & Interventions & 7.27 $\pm$ 0.58 & \textbf{87.50 $\pm$ 5.43} & \textbf{99.17 $\pm$ 1.02} & \textbf{100.00 $\pm$ 1.18} \\
    & Acoustic & \textbf{4.50 $\pm$ 0.73} & 59.58 $\pm$ 12.54 & 94.58 $\pm$ 2.12 & 98.75 $\pm$ 1.02 \\
     \midrule
    \multirow{3}{*}{UA Ensemble}
    & Disfluency & 6.29 $\pm$ 0.81 & 82.91 $\pm$ 6.37 & 97.91 $\pm$ 1.31 & 100.00 $\pm$ 0.00  \\
    & Interventions & \textbf{5.46 $\pm$ 1.57} & 73.75 $\pm$ 14.47 & 93.33 $\pm$ 5.17 & 97.91 $\pm$ 1.86 \\
    & Acoustic & 5.31 $\pm$ 1.30 & \textbf{75.41 $\pm$ 11.21} & \textbf{96.25 $\pm$ 3.06} & \textbf{99.16 $\pm$ 1.02} \\
     \midrule
    \multirow{3}{*}{UA Ensemble (weighted)}   
    & Disfluency & 6.29 $\pm$ 0.81 & 83.33 $\pm$ 6.58 & 97.91 $\pm$ 1.31 & 100.00 $\pm$ 0.00 \\
    & Interventions & \textbf{5.46 $\pm$ 1.57} & 76.25 $\pm$ 13.85 & 92.50 $\pm$ 5.98 & 96.66 $\pm$ 3.11 \\
    & Acoustic & 5.31 $\pm$ 1.30 & \textbf{75.83 $\pm$ 10.59} & \textbf{95.00 $\pm$ 3.86} & \textbf{99.16 $\pm$ 1.02} \\
    \bottomrule
  \end{tabular}
\label{tab:res_ad_mpiw}
\end{center}
\end{table*}

\begin{figure*}[htbp] 
\centering 
  \begin{minipage}{.32\textwidth}
  \centering
  \includegraphics[width=0.9\linewidth]{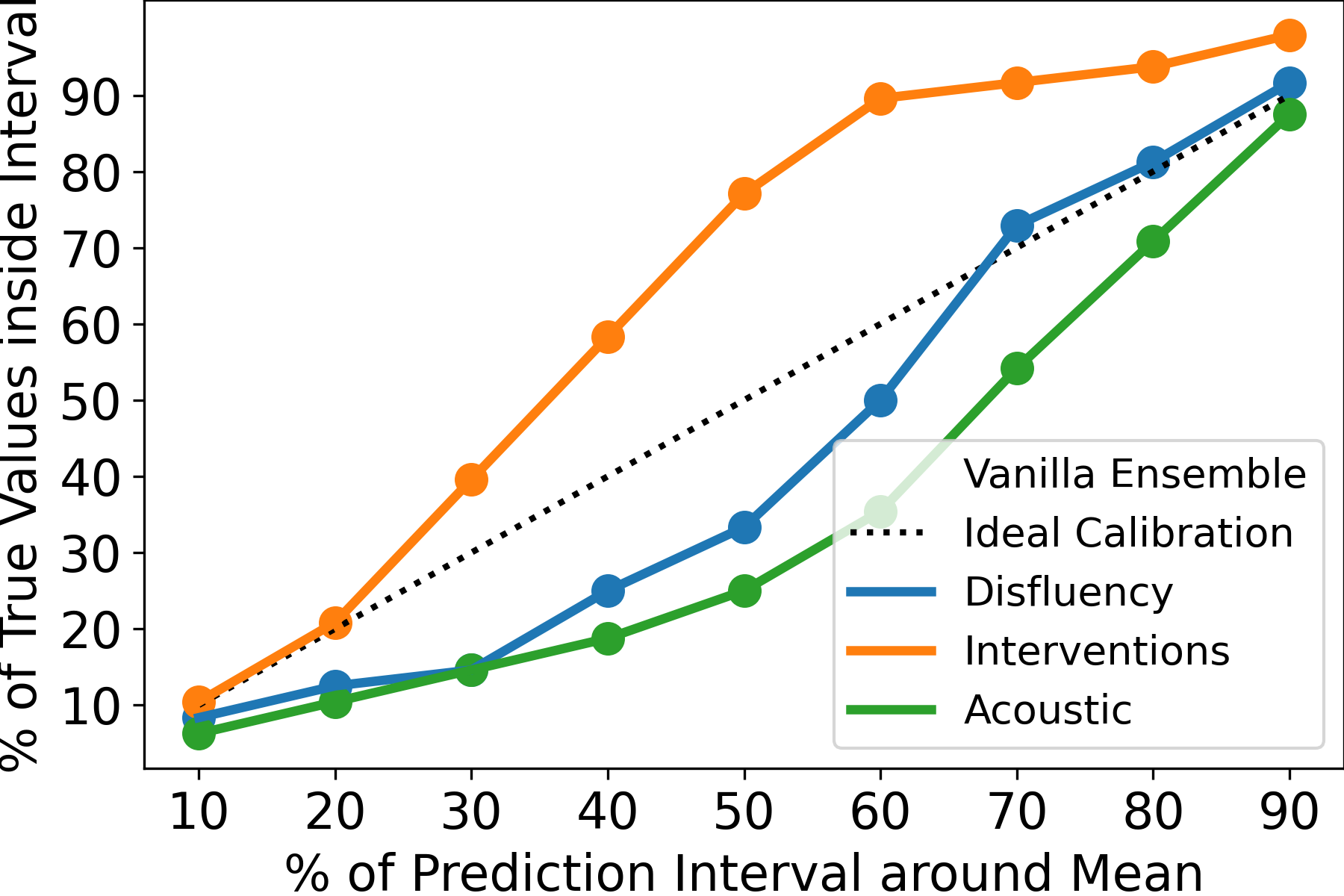}
\end{minipage}%
~~
\begin{minipage}{.32\textwidth}
\centering
  \includegraphics[width=0.9\linewidth]{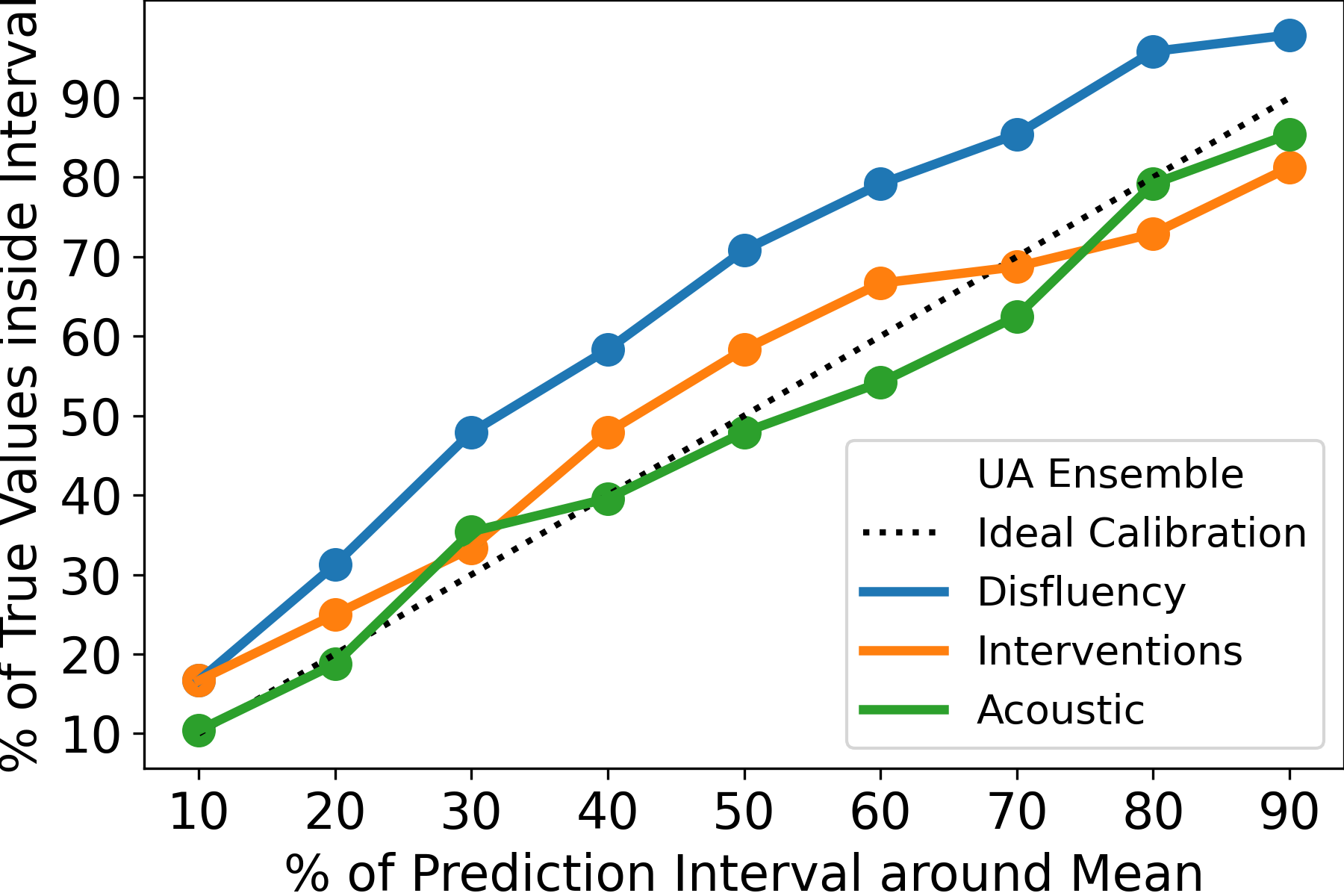}
\end{minipage}%
~~
\begin{minipage}{.32\textwidth}
\centering
  \includegraphics[width=0.9\linewidth]{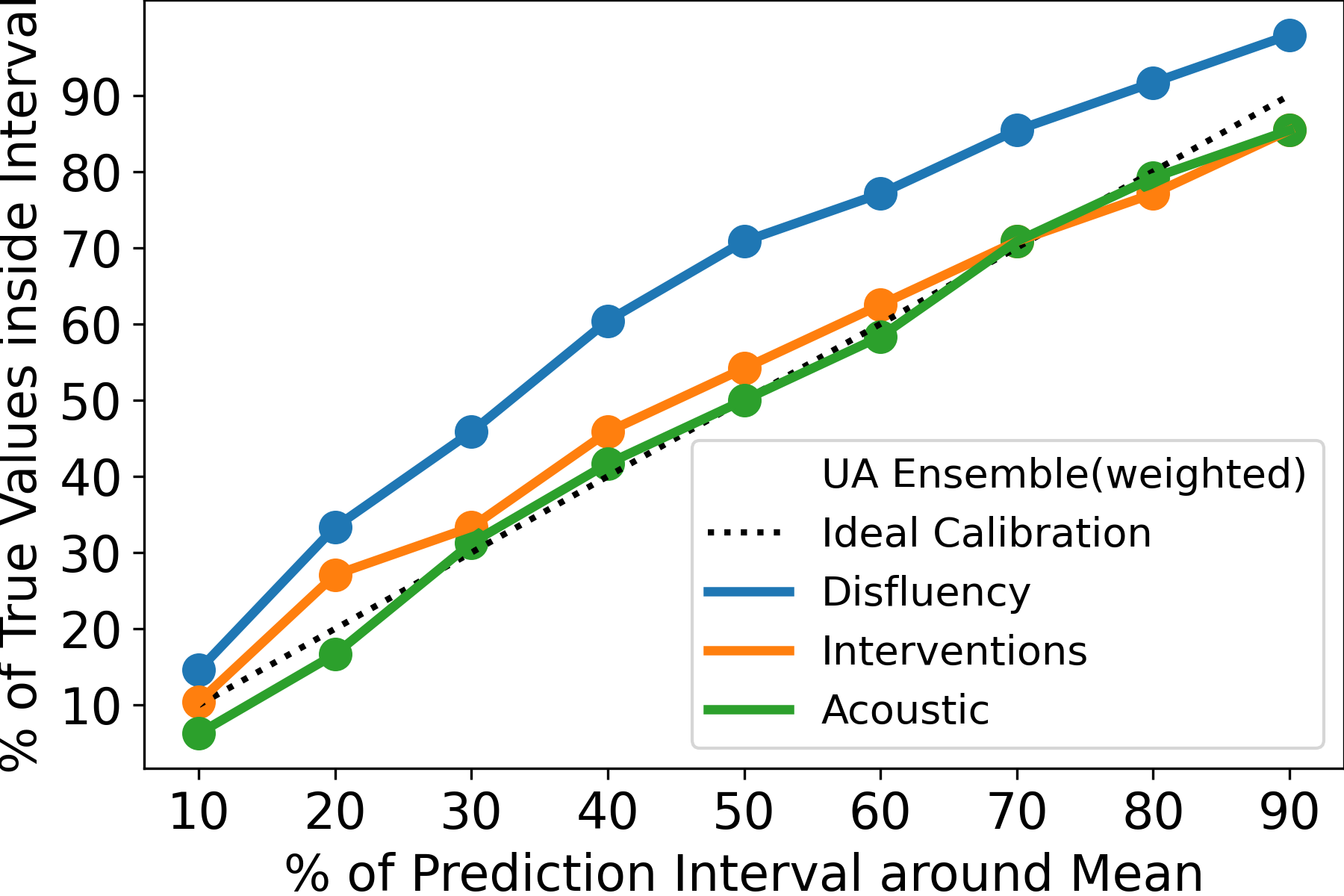}
\end{minipage}%
\\
\vspace{4mm}
  \begin{minipage}{.32\textwidth}
  \centering
  \includegraphics[width=0.9\linewidth]{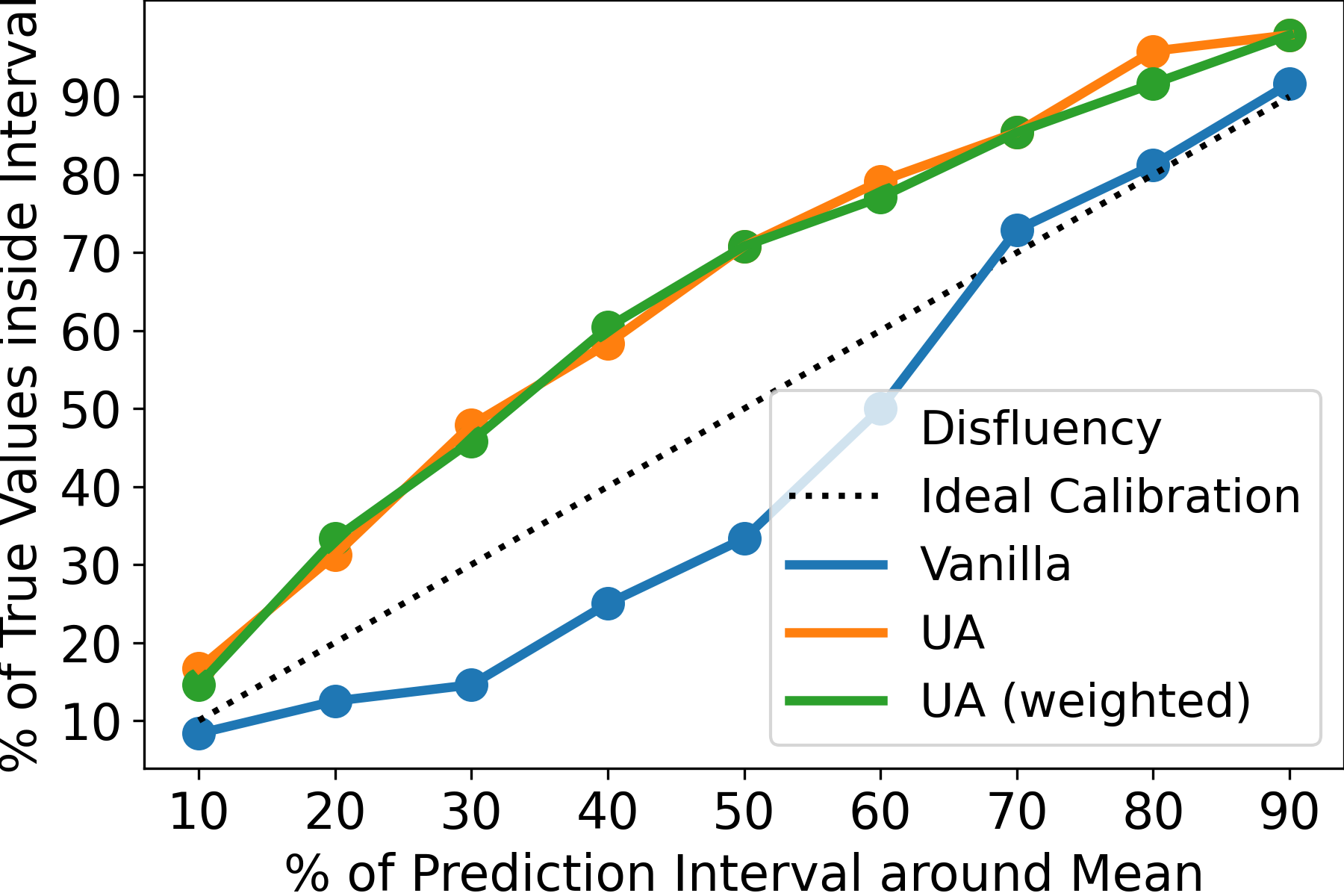}
\end{minipage}%
~~
\begin{minipage}{.32\textwidth}
\centering
  \includegraphics[width=0.9\linewidth]{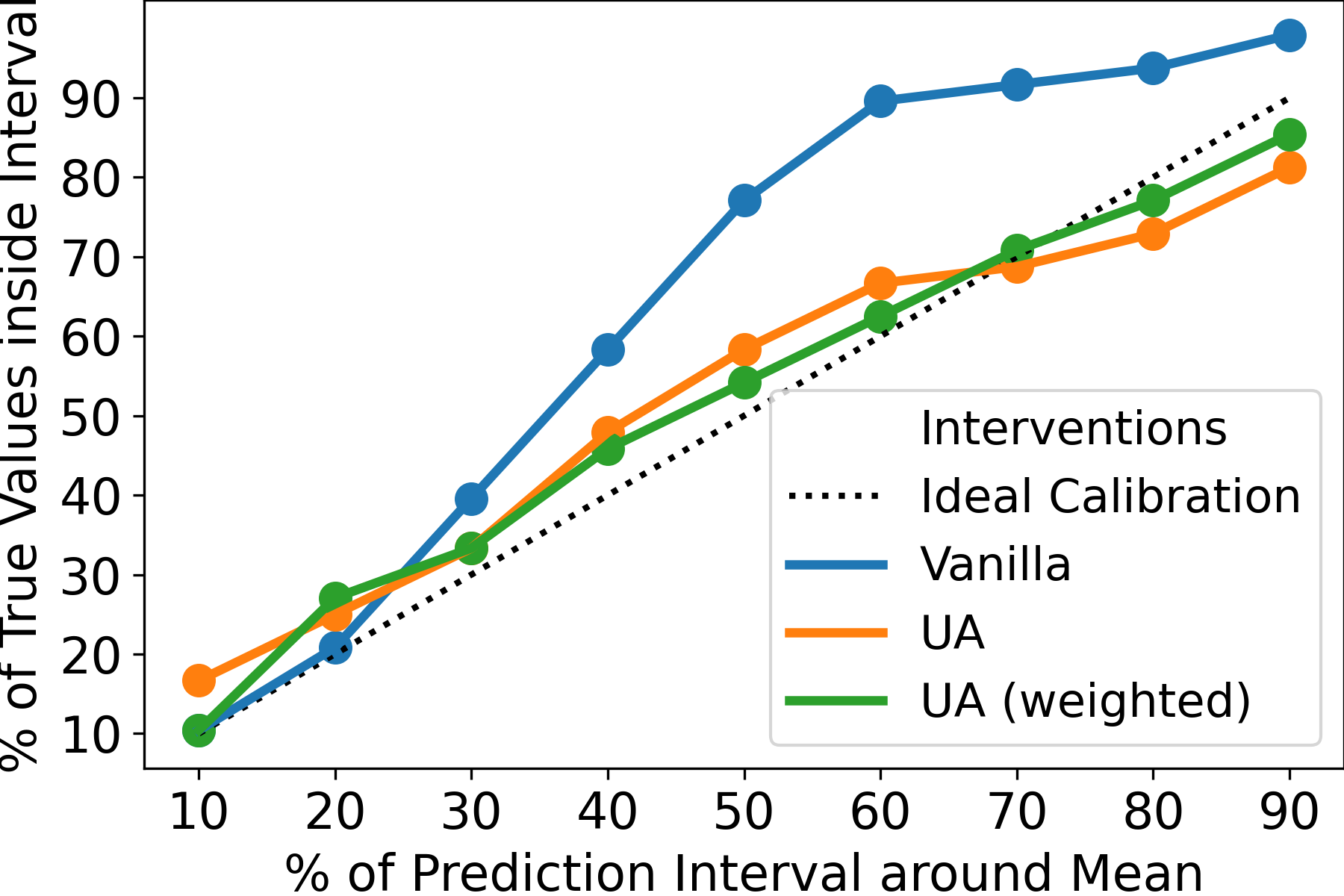}
\end{minipage}%
~~
\begin{minipage}{.32\textwidth}
\centering
  \includegraphics[width=0.9\linewidth]{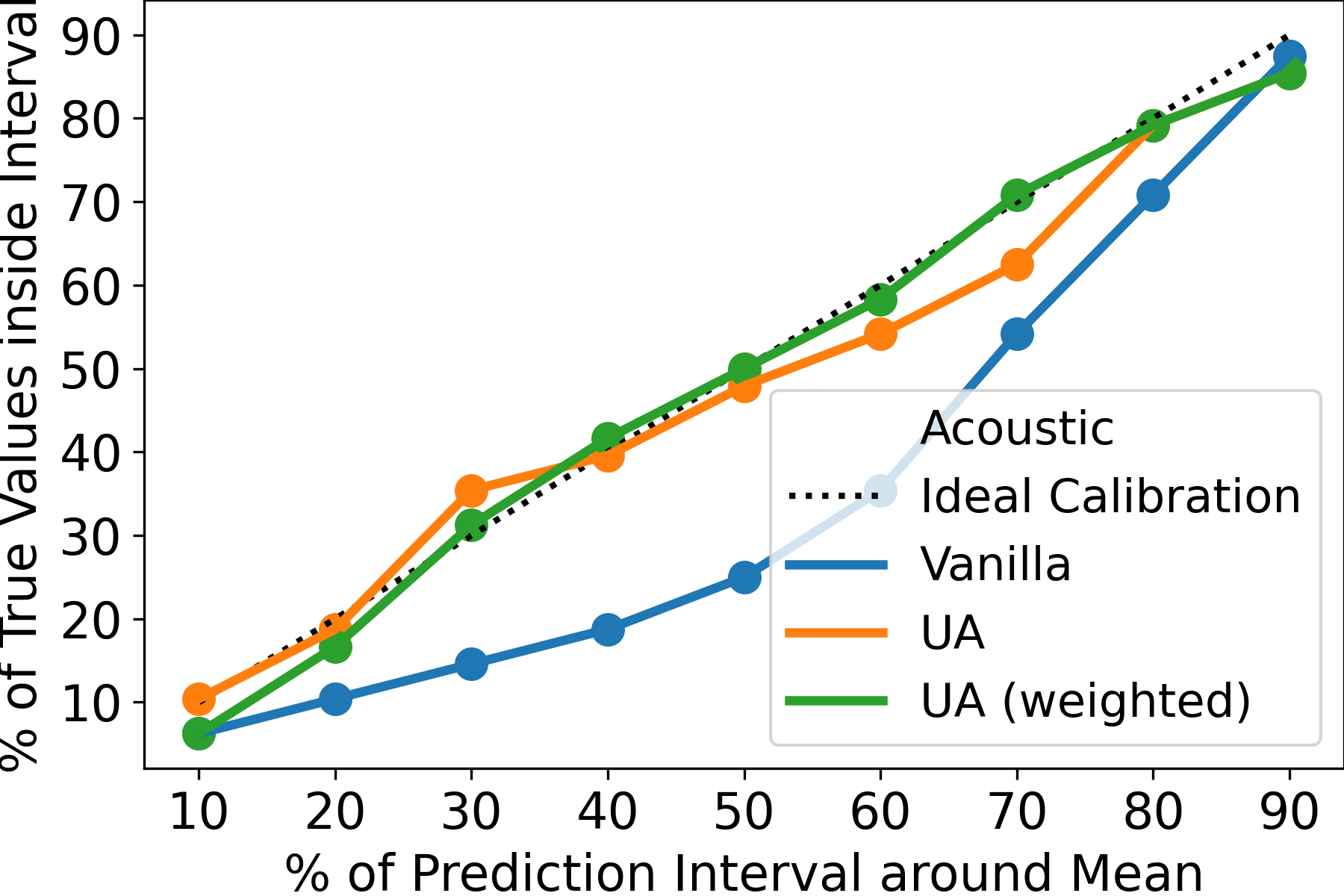}
\end{minipage}%

  \caption{Calibration curves for the ensemble techniques on the ADReSS dataset. The two rows of plots use the same data to just visualize the comparison differently (ensemble-wise and modality-wise respectively). The plots show that the boosted modalities are better-calibrated in case of uncertainty-aware boosting (Section~\ref{sec:ad_results}).}\label{fig:ad_empiricalplots}
\end{figure*}

We also analyse our approach on the Mean Prediction Interval Width (MPIW) and Prediction Interval Coverage Probability (PICP), two widely used metrics for evaluating uncertainty in regression. PICP is the percentage of the times the prediction interval contains the actual regression value, while MPIW is the average size of all prediction intervals. Pearce et al. \cite{pearce2018pi} discusses that high-quality prediction intervals should be as narrow as possible, whilst capturing some specified proportion of data points. Accordingly, it is desirable to have low MPIW values and PICP $\ge (1-\alpha)$, a common choice of $\alpha$ being $0.05$.

Table~\ref{tab:res_ad_mpiw} shows the 5-times repeated test set results on MPIW and PICP metrics with a comparison between the vanilla ensemble and the uncertainty-aware ensembles. It is important to note that the comparison holds value for the results corresponding to the modalities that have been boosted (in this case, the Interventions and Acoustic modalities). UA ensemble and its variation UA ensemble (weighted) observe reduced MPIW compared to vanilla ensemble for the Interventions modality. While the vanilla ensemble observes a reduced MPIW for the Acoustic modality, the uncertainty-aware ensembles observe a gradual decrease in the MPIW values while sequentially boosting across the modalities. The uncertainty-aware boosting thus results in tighter bounds for the confidence intervals, along with higher PICP values, and high quality prediction intervals as desired \cite{pearce2018pi}.

We further use the 65-95-99.7 rule (also called the empirical rule) to obtain calibration curves for a comprehensive analysis of calibration \cite{lakshminarayanan2017simple, sarawgi2020unified}. To plot these curves, we first compute the $x\%$ prediction interval for each data point under evaluation based on Gaussian quantiles using the prediction value and variance. We then calculate the fraction of data points under evaluation with true values that fall within this prediction interval. For a well-calibrated model, the observed fraction should be close to the $x\%$ calculated earlier. To see how our models perform in this setting, we sweep from $x = 10\%$ to $x = 90\%$ in steps of 10. A line lying close to the line ($y=x$) would indicate a well-calibrated model.

Fig.~\ref{fig:ad_empiricalplots} shows the calibration curves of the three modalities in case of vanilla ensemble, UA ensemble, and UA ensemble (weighted). We observe that in the case of UA ensemble and UA ensemble (weighted), the Interventions and Acoustic modalities become better-calibrated compared to the Disfluency modality. However in case of vanilla ensemble, the calibration of the Interventions and Acoustic modalities become worse when compared to the Disfluency modality. Consequently, it follows that the boosted modalities are better-calibrated in case of uncertainty-aware boosting, as desired in real-world settings. This highlights the significance of introducing the notion of uncertainty-awareness in ensembles to obtain better-calibrated models.

\subsubsection{Parkinson's Disease}\label{sec:pd_results}
\begin{figure*}[htbp] 
\centering 
  \begin{minipage}{.32\textwidth}
  \centering
  \includegraphics[width=0.9\linewidth]{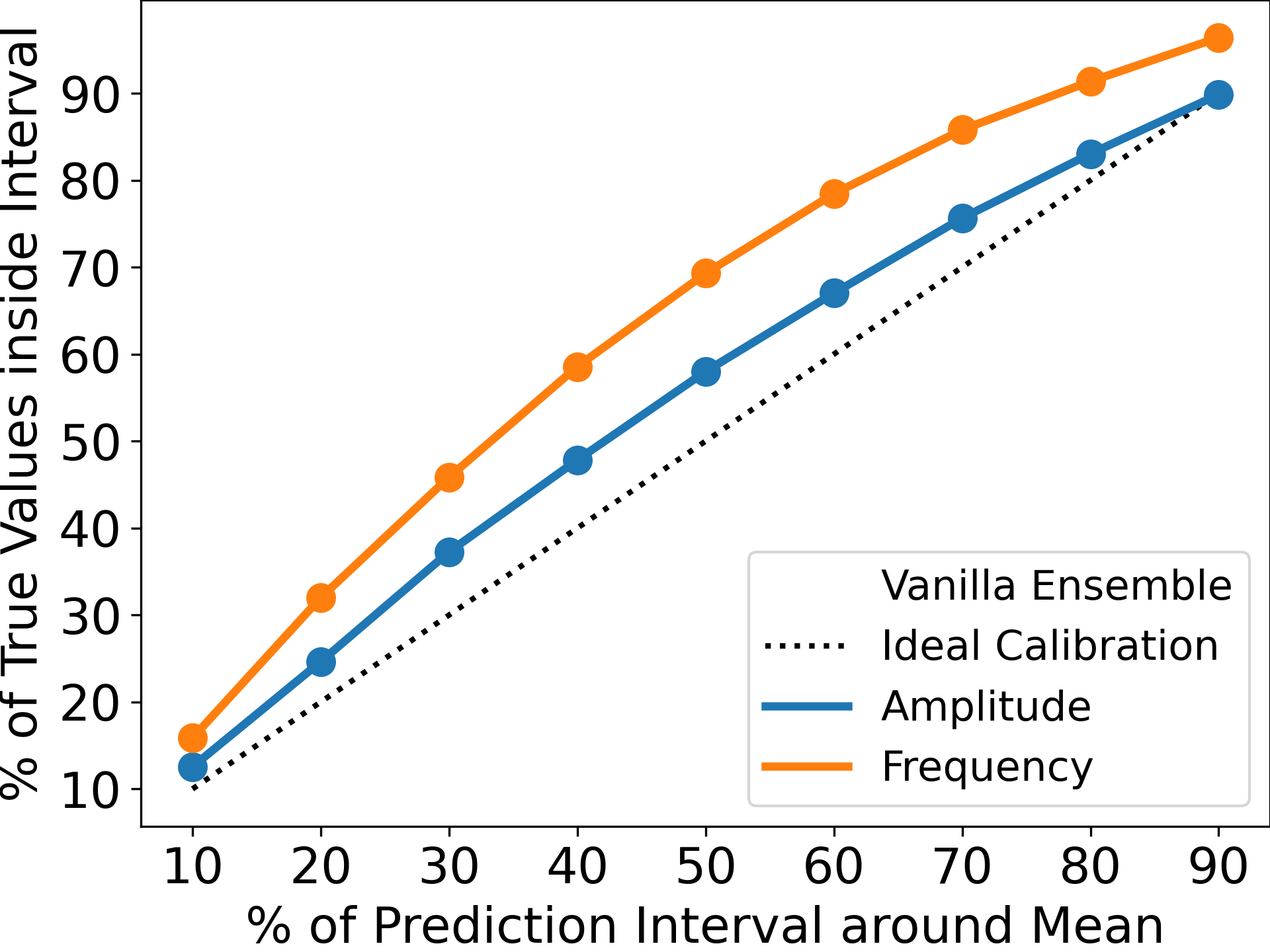}
\end{minipage}%
~~
\begin{minipage}{.32\textwidth}
\centering
  \includegraphics[width=0.9\linewidth]{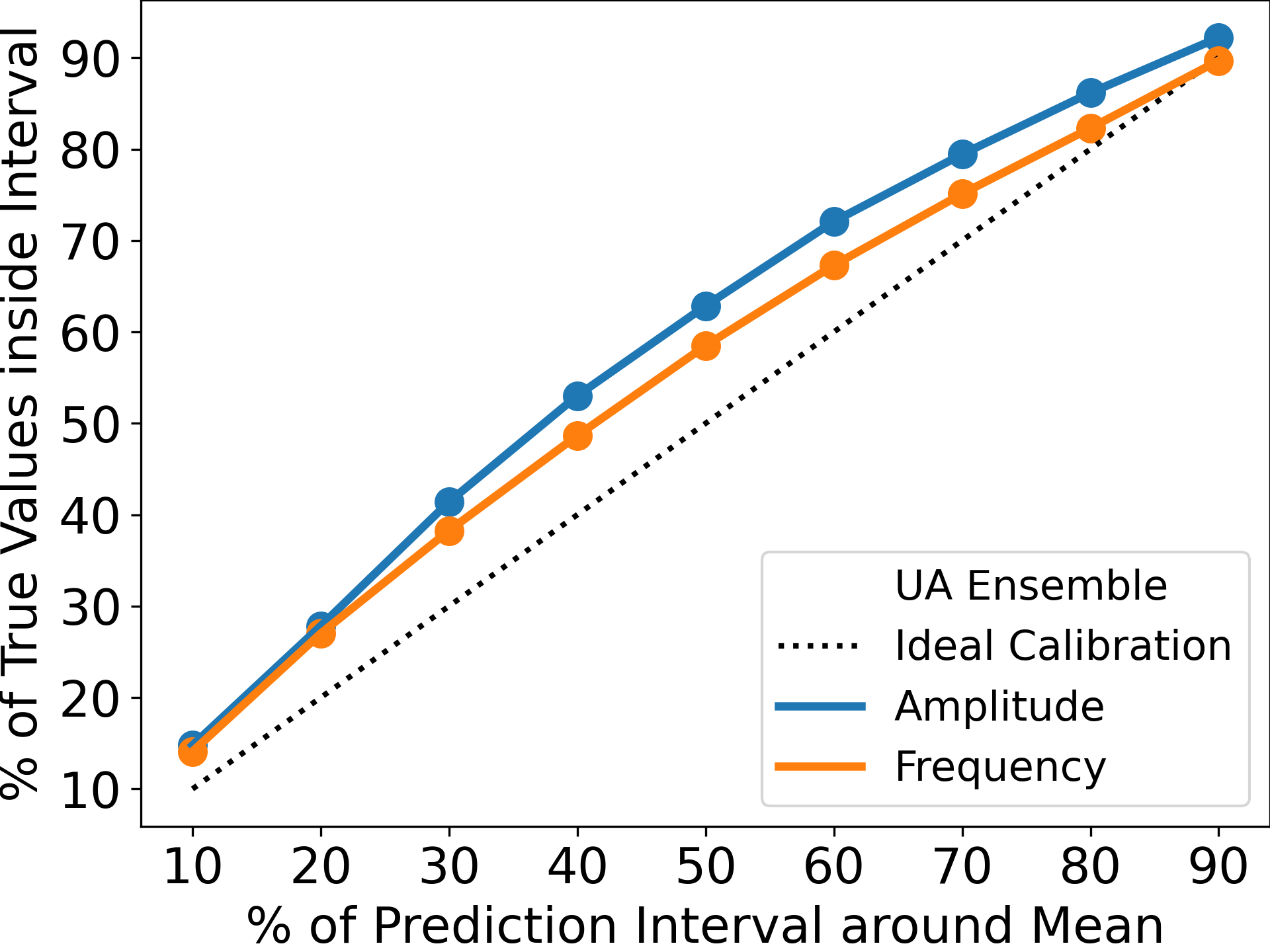}
\end{minipage}%
~~
\begin{minipage}{.32\textwidth}
\centering
  \includegraphics[width=0.9\linewidth]{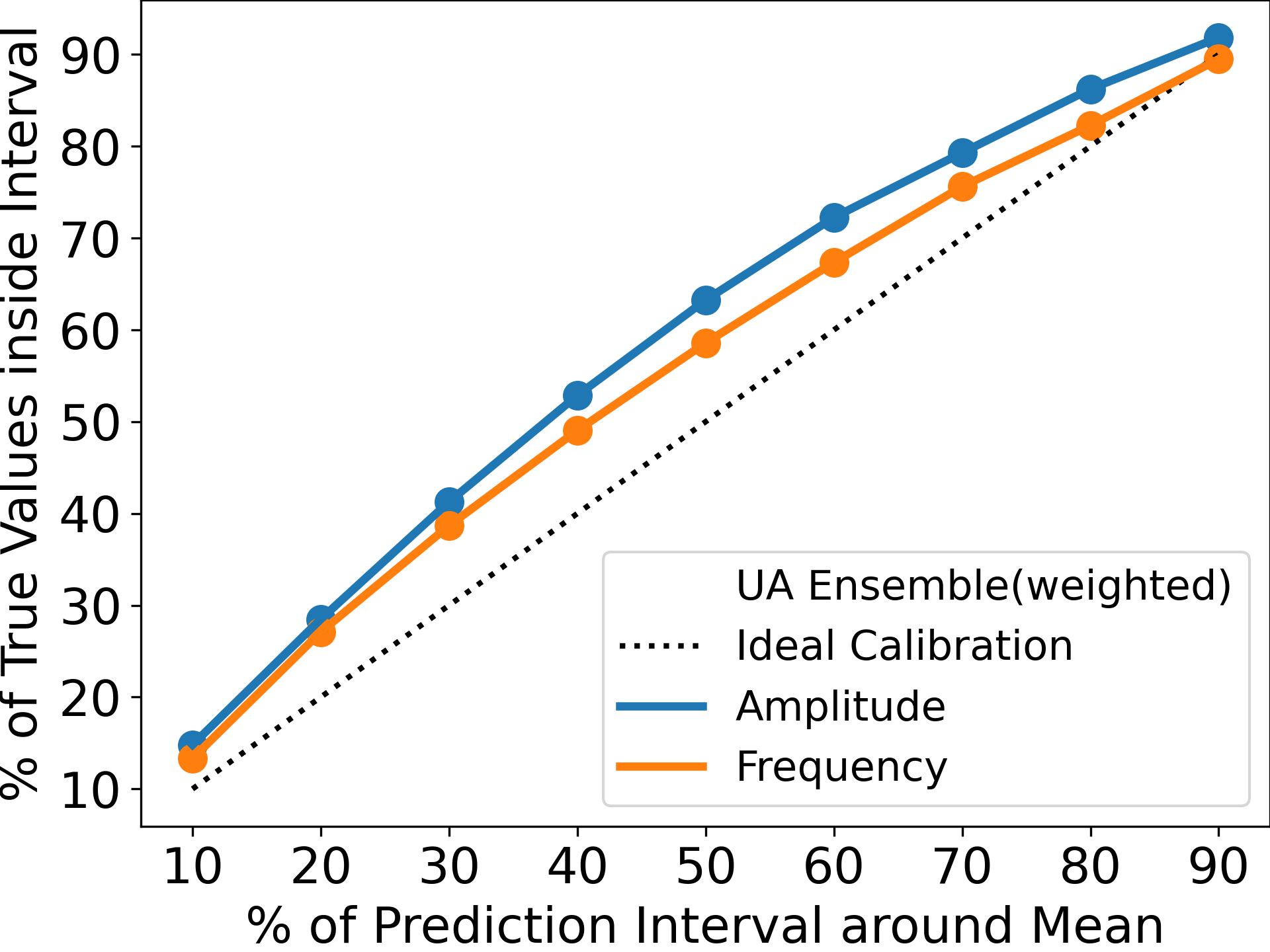}
\end{minipage}%
\\
\vspace{4mm}
  \begin{minipage}{.32\textwidth}
  \centering
  \includegraphics[width=0.9\linewidth]{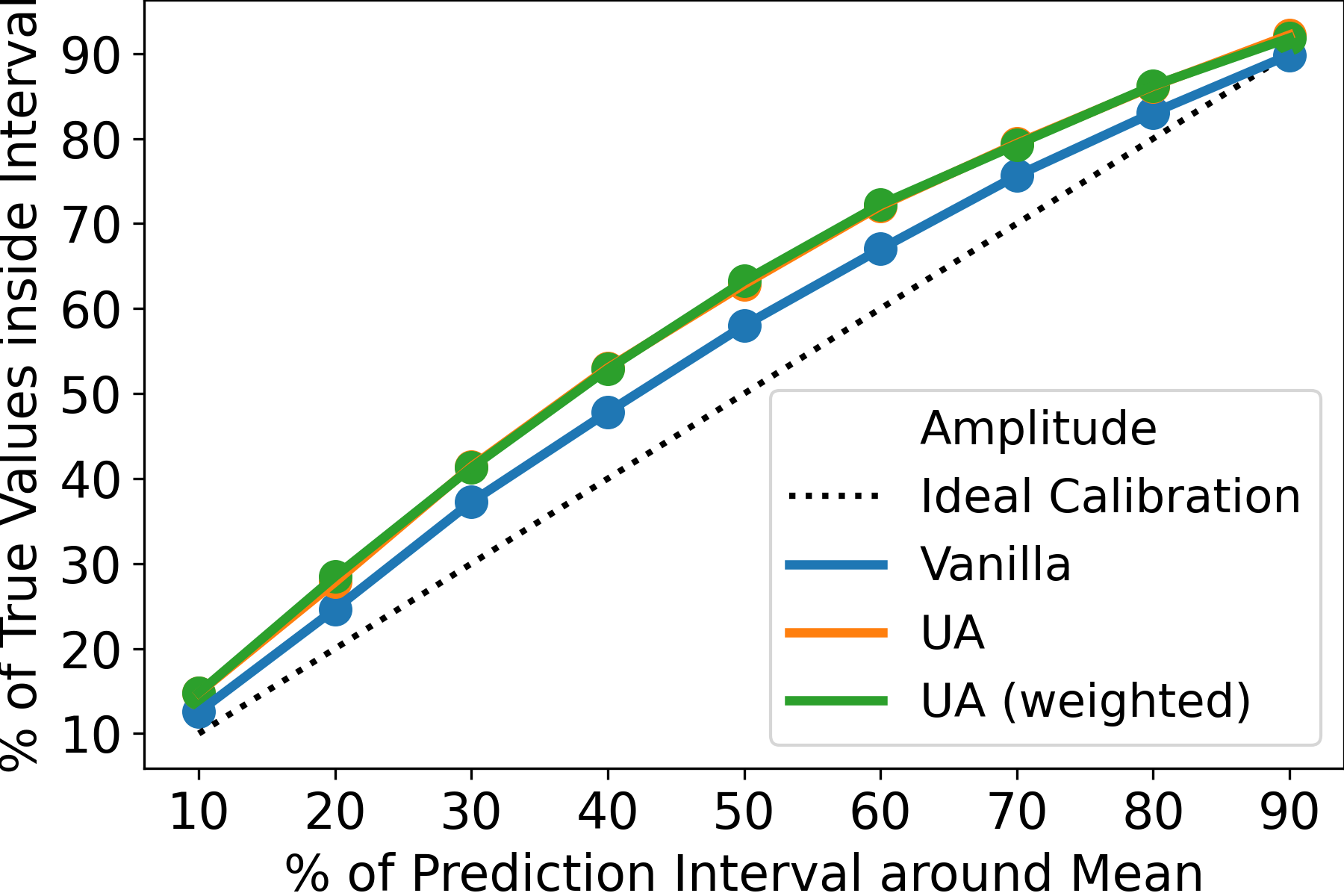}
\end{minipage}%
~~
\begin{minipage}{.32\textwidth}
\centering
  \includegraphics[width=0.9\linewidth]{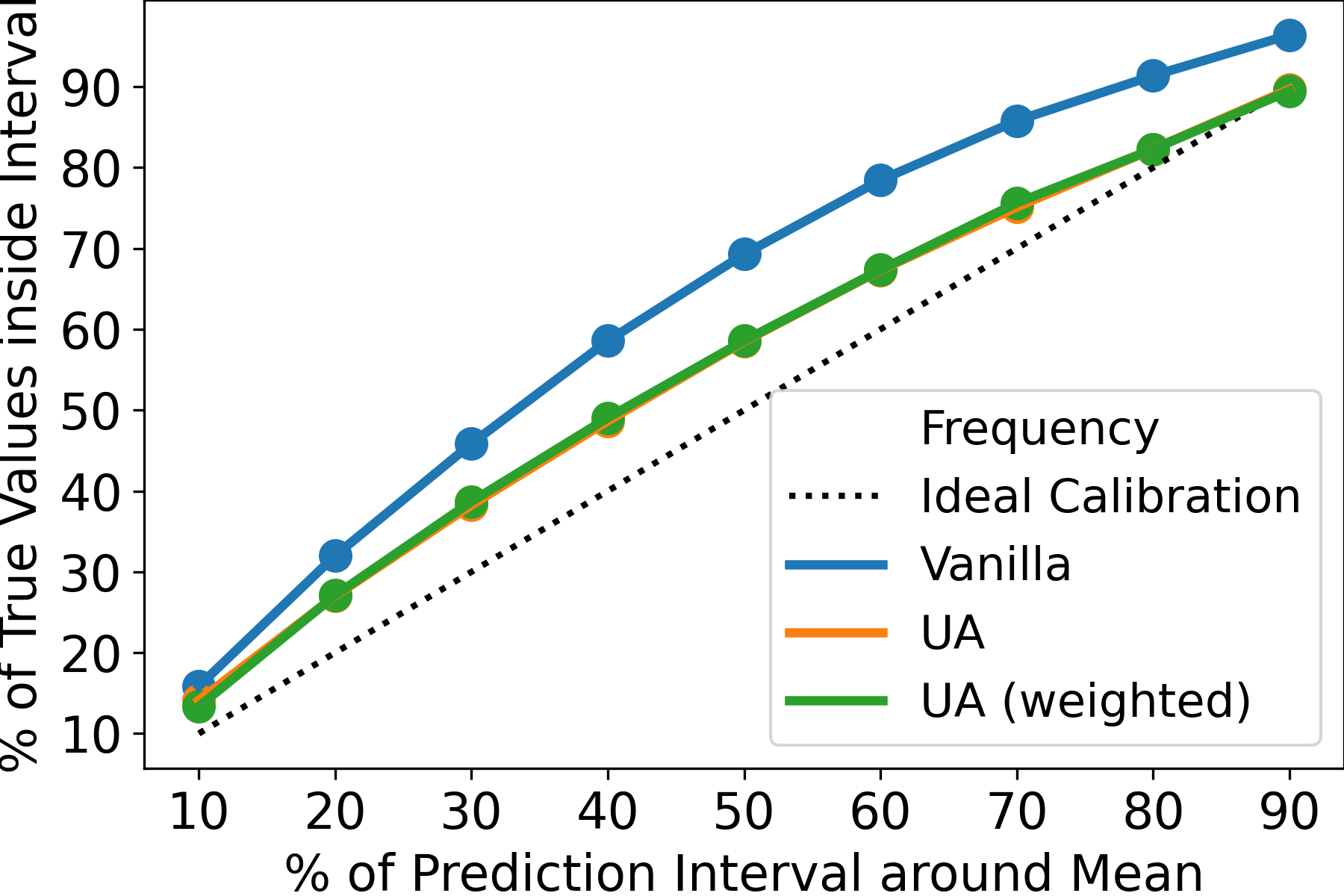}
\end{minipage}%

  \caption{Calibration curves for the ensemble techniques on the Parkinson's Telemonitoring dataset. The two rows of plots use the same data to just visualize the comparison differently (ensemble-wise and modality-wise respectively). Also in the second row, the green and orange plots almost overlap (and hence, might not be clearly visible). The plots show that the boosted modalities are better-calibrated in case of uncertainty-aware boosting (Section~\ref{sec:pd_results}).}\label{fig:pd_empiricalplots}
\end{figure*}
Since there is no standardised train-test split available, we perform a $5$-fold cross validation and report the mean and variance of the RMSE results across the five folds. We first evaluate each of the modalities (i.e. base learners) individually and then compare them with the vanilla and uncertainty-aware ensembles. Again, the order of sequential boosting for the propagation of the uncertainties is chosen in the order of the cross-validation performance of the individual modalities. We observe that the uncertainty-aware ensembles perform better than the vanilla ensemble and the individual modalities (Table~\ref{tab:res_parkinson}).

\begin{table}[htbp]
\caption{Comparison of individual modalities i.e. base learners and ensemble methods on 5-fold cross validation results of the Parkinson's Telemonitoring dataset.}
\begin{center}
\begin{tabular}{lc}
  \toprule
  \bfseries Model & \bfseries RMSE\\
  \midrule
  Amplitude & 3.21 $\pm$ 0.06\\
  Frequency & 3.32 $\pm$ 0.10\\
  Vanilla Ensemble & 3.18 $\pm$ 0.05\\
  \textbf{UA Ensemble} & \textbf{3.04 $\pm$ 0.04}\\
  \textbf{UA Ensemble (weighted)} & \textbf{3.05 $\pm$ 0.05}\\
  \bottomrule
\end{tabular}
\label{tab:res_parkinson}
\end{center}
\end{table}


Table~\ref{tab:res_parkinson2} shows the 5-fold cross-validation results on MPIW and PICP metrics with a comparison between the vanilla ensemble and the uncertainty-aware ensembles. It is important to note that the comparison holds value for the results corresponding to the modalities that have been boosted (in this case, the Frequency modality). UA ensemble and its variation UA ensemble (weighted) observe reduced MPIW compared to vanilla ensemble, indicating that the uncertainty-aware boosting results in tighter bounds for the confidence intervals, along with higher PICP values, and high quality prediction intervals as desired \cite{pearce2018pi}.

Fig.~\ref{fig:pd_empiricalplots} shows the calibration curves of the two modalities in case of vanilla ensemble, UA ensemble, and UA ensemble (weighted). We observe that in the case of UA ensemble and UA ensemble (weighted), the Frequency modality becomes better-calibrated compared to the Amplitude modality. However in case of vanilla ensemble, the calibration of the Frequency modality becomes worse when compared to the Amplitude modality. Consequently, as also observed with the ADReSS dataset, it follows that the boosted modality is better-calibrated in case of uncertainty-aware boosting, as desired in real-world settings. This again highlights the significance of introducing the notion of uncertainty-awareness in ensembles to obtain better-calibrated models.



\begin{table*}[htbp]
\caption{5-fold cross-validation results of Mean Prediction Interval Width (MPIW) and Prediction Interval Coverage Probability (PICP) for the ensemble techniques on the Parkinson's Telemonitoring dataset. We report PICP results with the prediction interval ($\Delta$) equal to 1, 2, and 3 times the standard deviation (i.e. $1\sigma$, $2\sigma$, and $3\sigma$). The uncertainty-aware boosting results in tighter bounds for the confidence intervals, along with higher PICP values, and high quality prediction intervals as desired (Section~\ref{sec:pd_results}).}
\begin{center}
\begin{tabular}{lccccc}
    \toprule
    \multirow{2}{*}{Model}     & \multirow{2}{*}{Modality}   & 
    \multirow{2}{*}{MPIW}   & 
    \multicolumn{3}{c}{PICP ($\%$)} \\
    \cmidrule(r){4-6} 
    &   &   & $\Delta = 1\sigma$ & $\Delta = 2\sigma$ & $\Delta = 3\sigma$\\
    \toprule
    \multirow{2}{*}{Vanilla Ensemble}   
    & Amplitude & 6.79 $\pm$ 1.28 & 84.56 $\pm$ 1.46 & 98.51 $\pm$ 0.58 & 99.89 $\pm$ 0.12 \\
    & Frequency & 8.69 $\pm$ 0.59 & 74.17 $\pm$ 8.25 & 94.28 $\pm$ 3.37 & 98.60 $\pm$ 1.18 \\
     \midrule
    \multirow{2}{*}{UA Ensemble}   
    & Amplitude & 6.50 $\pm$ 1.76 & 74.09 $\pm$ 9.15 & 93.70 $\pm$ 4.11 & 98.23 $\pm$ 1.47 \\
    & Frequency & \textbf{6.91 $\pm$ 0.85} & \textbf{77.90 $\pm$ 5.28} & \textbf{95.64 $\pm$ 2.40} & \textbf{99.33 $\pm$ 0.51} \\
     \midrule
    \multirow{2}{*}{UA Ensemble (weighted)}   
    & Amplitude & 6.50 $\pm$ 1.76 & 74.24 $\pm$ 8.59 & 93.71 $\pm$ 4.13 & 97.97 $\pm$ 1.66 \\
    & Frequency & \textbf{6.91 $\pm$ 0.85} & \textbf{77.65 $\pm$ 5.67} & \textbf{95.45 $\pm$ 2.56} & \textbf{99.18 $\pm$ 0.70} \\
    \bottomrule
  \end{tabular}
\label{tab:res_parkinson2}
\end{center}
\end{table*}

\section{Discussion and Future Work}
\label{sec:discussion}

We proposed an uncertainty-aware boosted ensembling method in multi-modal settings. Such an ensemble improves the performance when compared to individual modalities and ensembles boosted using loss values. By focusing more on data points with higher uncertainty, through uncertainty-weighting of the loss function (UA Ensemble) and the predictions as well (UA Ensemble (weighted)), we showed how our ensemble outperforms the results of state-of-the-art methods. More importantly, our discussion in Section~\ref{sec:introoo} highlight how such an ensemble system can help design a more robust learner by having the base learners pay more attention to uncertain prediction corresponding to noisy data modalities. Our experiments showed that the propagation of the uncertainty sequentially through the base learners of every modality aids the multi-modal system to decrease the overall entropy in the system, making it more reliable when compared to vanilla ensembles. Additionally, the modalities indeed become well calibrated along with high quality prediction intervals when boosted using uncertainty values, rather than loss values.

Such characteristics are significantly desired in real-world settings where data tends to exist in multiple modalities together. Understanding what a machine learning model does not know is crucial in safety-critical applications. Access to such information helps with designing a more reliable and aware decision-making system \cite{amodei2016concrete, varshney2017safety, kumar2019verified, thiagarajan2020building, gal2016uncertainty}. Furthermore, the availability of predictive uncertainties corresponding to each modality adds a level of transparency to the machine learning system. This can assist the user in making more informed decisions, thereby nurturing the synergy between humans and AI.

There are a lot of interesting possible future research directions to this work. One could definitely expand the proposed method itself to account for uncertainty values, as well as loss values, while boosting the base learners. Our current experiments make use of speech and text data with neural networks and random forests as the base learners. This can be extended to other forms of machine learning systems, making use of other Bayesian and non-Bayesian uncertainty estimation techniques and data modalities. Additionally, we encourage the community to further evaluate such techniques in other safety-critical tasks and applications, as well as assess the longitudinal performance and attributes of these systems, especially in the presence of noisy data and/or when the observed data distribution tends to shift over time and eventually becomes very different. This also opens up avenues to potentially design adaptive systems which could actively learn from the uncertainty estimates at deployment time.\\


\bibliographystyle{IEEEtran}
\bibliography{ref}

\end{document}